%% file: main.tex
\PassOptionsToPackage{table,xcdraw}{xcolor}
\documentclass[sigconf, 10pt, nonacm, natbib=false]{acmart}

\usepackage{color}
\usepackage{xspace}
\usepackage{subcaption}
\usepackage{graphicx}

\usepackage[normalem]{ulem}
\useunder{\uline}{\ul}{}

\AtBeginDocument{%
  \providecommand\BibTeX{{%
    \normalfont B\kern-0.5em{\scshape i\kern-0.25em b}\kern-0.8em\TeX}}}

\setcopyright{acmcopyright}
\copyrightyear{2018}
\acmYear{2018}
\acmDOI{XXXXXXX.XXXXXXX}

\acmConference[Conference acronym 'XX]{Make sure to enter the correct
  conference title from your rights confirmation email}{June 03--05,
  2018}{Woodstock, NY}
\acmISBN{978-1-4503-XXXX-X/18/06}

\usepackage[%
  style=acmnumeric, 
  backend=biber,
  sorting=none,
  maxbibnames=1,
  firstinits=true
]{biblatex}
\begin{document}

\newcommand{\shortname}{\textit{ML Mule}\xspace}

\title{{\shortname: Mobile-Driven\\Context-Aware Collaborative Learning}}

\author{Haoxiang Yu}
\email{hxyu@utexas.edu}
\orcid{0000-0002-3518-946X}
\affiliation{%
  \institution{University of Texas at Austin}
  \city{Austin}
  \state{Texas}
  \country{USA}
}

\author{Javier Berrocal}
\email{jberolm@unex.es}
\orcid{0000-0002-1007-2134}
\affiliation{%
  \institution{University of Extremadura}
  \city{Badajoz}
  \state{Badajoz}
  \country{Spain}
}

\author{Christine Julien}
\email{christinejulien@vt.edu}
\orcid{0000-0002-4131-4642}
\affiliation{%
  \institution{Virginia Tech}
  \city{Blacksburg}
  \state{Virginia}
  \country{USA}
}

\begin{abstract}
Artificial intelligence has been integrated into nearly every aspect of daily life, powering applications from object detection with computer vision to large language models for writing emails and compact models for use in smart homes. 
These machine learning models at times cater to the needs of individual users but are often detached from them, as they are typically stored and processed in centralized data centers.
This centralized approach raises privacy concerns, incurs high infrastructure costs, and struggles to provide real time, personalized experiences. Federated and fully decentralized learning methods have been proposed to address these issues, but they still depend on centralized servers or face slow convergence due to communication constraints.
We propose \shortname, an approach that utilizes individual mobile devices as ``mules'' to train and transport model snapshots as the mules move through physical spaces, sharing these models with the physical ``spaces'' the mules inhabit. This method implicitly forms affinity groups among devices associated with users who share particular spaces, enabling collaborative model evolution and protecting users' privacy. Our approach addresses several major shortcomings of traditional, federated, and fully decentralized learning systems. \shortname represents a new class of machine learning methods that are more robust, distributed, and personalized, bringing the field closer to realizing the original vision of intelligent, adaptive, and genuinely context-aware smart environments. Our results show that \shortname converges faster and achieves higher model accuracy compared to other existing methods.
\end{abstract}



\maketitle

\section{Introduction}
\input{sections/1_intro}
\section{Related Work}
\input{sections/2_related}
\section{System Design}
\input{sections/3_system}
\section{Evaluation}
\input{sections/4_evaluation}

\section{Conclusion}
\input{sections/6_conclusion}

\printbibliography

\end{document}

%% file: sections/1_intro.tex
AI-driven technologies are transforming modern life, streamlining processes, and embedding themselves into everyday routines. Individuals interact with machine learning models constantly—whether turning on lights with a virtual assistant in the morning, composing emails using ChatGPT at work, tracking calories burned with a smartwatch, or letting a smart thermostat adjust the temperature before sleep. These interactions demonstrate how deeply machine learning models are integrated into our lives, yet the way these models operate often falls short of realizing the vision of truly ``smart'' environments.

\begin{figure}[!t]
  \centering
  \includegraphics[width=0.45\textwidth]{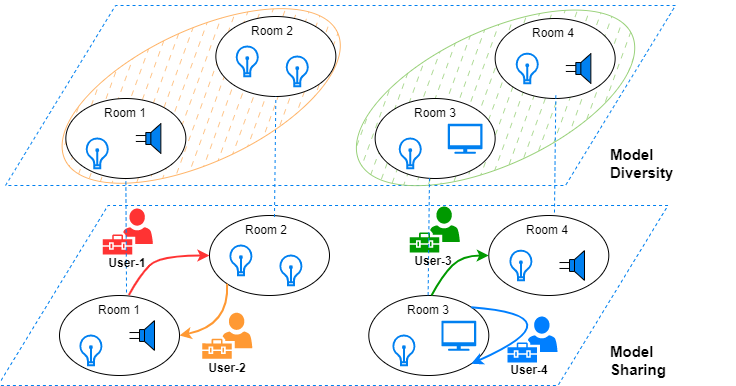}
  \caption{Example of \shortname sharing process}
  \Description[Overview of a system illustrating rooms, devices, and user interactions]{The diagram depicts two layers. The top layer represents four rooms (Room 1, Room 2, Room 3, and Room 4), each containing devices and showing their relationships. Room 1 contains a light bulb and a speaker, and it shares a similar characteristic with Room 2, which contains two light bulbs, indicated by an orange dotted oval. Room 3 has a light bulb and a computer, and it shares a similar characteristic with Room 4, which contains a light bulb and a speaker, as shown by a green dotted oval. The bottom layer highlights user interactions. User-1 (red) and User-2 (orange) both interact with Room 1 and Room 2. Similarly, User-3 (green) and User-4 (blue) interact with Room 3 and Room 4.}
  \label{fig:ml_mule} 
\end{figure}

Most machine learning models used in these applications are trained, stored, and executed on centralized servers far removed from the end users they support and the smart environments they control~\cite{lei2017insecurity}. While this remote infrastructure enables powerful computation and scalability, it also introduces significant challenges. Centralized systems pose risks to user privacy, have high infrastructure costs, and struggle to dynamically adapt to the needs of individual users~\cite{cho2020will}. Moreover, they fail to provide the contextual intelligence and adaptive behavior that are key features of a genuinely smart environment. The rapid advancement of generative AI, which often relies on even larger and more complex models, further highlights these limitations, as the need to balance privacy, scalability, and real-time adaptation becomes increasingly critical~\cite{huang2024federated}. In an ideal scenario, machine learning models would remain closer to users, learning and evolving directly from their unique experiences and interactions.

Classic federated learning~\cite{hard2018federated} requires collaborators to be coupled in time (i.e., all connected to the Internet simultaneously but perhaps in disparate locations), while decentralized approaches~\cite{hegedHus2019gossip, 9439130}, which rely on local device-to-device communication, require collaborators to be coupled in both space and time. However, to the best of our knowledge, no existing approaches require coupling only in space while allowing collaborators to be decoupled in time. 

In this work, we propose a distributed machine learning approach, called \shortname, that utilizes the spatial mobility of users and their devices as they naturally transition between different physical spaces. Our novel insight is that even when devices are decoupled in time, their spatial coupling can allow them to learn from one another, thereby enhancing the influence of the spatial context on model performance. In the context of existing decentralized approaches, devices collaborate when they ``encounter'' one another, which is commonly defined as the devices appearing in the same space at the same time; however, with \shortname, devices only need to be in the same space, even if they are not present simultaneously. Such an approach is particularly appealing for a particular subset of applications in which the physical space has particular importance to what the model learns.

Our framework entails two key roles: (1)~fixed devices embedded in physical spaces, and (2)~mobile devices carried by users. When a mobile device enters a space with fixed devices, the fixed and mobile devices collaboratively train a model using locally acquired data. The trained model is then stored on both the fixed and mobile devices. As a user moves to a new space, their mobile device carries a snapshot of the model that they received from and trained alongside the previously encountered fixed device.
Upon arriving in a new space, the user's mobile device can share this carried model with a fixed device in the new environment. Similarly, a fixed device shares its local model snapshot with any newly arriving mobile device that enters the space.

Figure~\ref{fig:ml_mule} illustrates an example of \shortname in a smart home control setting. Fixed devices, such as smart lights and speakers, collaborate with mobile devices carried by users to train and exchange model snapshots. For instance, in this contrived example, User-1 (red) trains a model in Room 1, carries it to Room 2, and shares it with the fixed devices in Room 2. Similarly, User-2 (orange) and other users transport model snapshots between rooms, enabling fixed devices in different rooms to update
their models collaboratively. In this figure, User-1 and User-2 frequently enter Rooms 1 and 2, sharing similar characteristics. Likewise, User-3 (green) and User-4 (blue) interact between Rooms 3 and 4, sharing similar characteristics with each other but different from those of Users 1 and 2 in Rooms 1 and 2.

Our \shortname framework creates a dynamic and collaborative learning ecosystem in which mobile devices act as {\em mules}, a term inspired by its use in delay-tolerant networks, where mules refer to mobile agents that transport and exchange data between disconnected nodes~\cite{6702844}.
Similarly, in our proposed \shortname framework, mobile devices are
transporting and exchanging model updates between physical spaces. By implicitly forming affinity groups among devices that overlap by virtue of their shared spaces, our approach enables localized and context-aware learning through aggregation of models that incorporate contextual information. 
This collaborative learning mechanism presumes that users who share physical spaces are likely to exhibit similar characteristics, enabling the creation of more nuanced, personalized, and adaptive models that can be tailored to how these users use the spaces they inhabit. 

In employing \shortname,
a model is enhanced through human interaction---\shortname is not a wholesale replacement for all use cases suited to federated learning or decentralized learning. Instead, it is designed to address specific applications where this approach provides clear advantages, while other applications may continue to benefit from existing methods or from combinations of those methods with an \shortname inspired approach. The particular use cases for which \shortname is particularly fitting are those in which the space significantly influences the learning task. These include, for example, applications in smart environments, where devices in the space attempt to learn how to configure themselves to support the users in that space. Human activity recognition applications are another example --- different humans often perform the same or similar activities in the same space (e.g., a gym, a restaurant, or a movie theater).

Our work represents a step forward in achieving distributed machine learning systems that are robust, realistic, privacy-preserving, and tailored to both human-centric and space-centric
needs. In summary, our main contributions are: %
\begin{enumerate}
    \item We introduce \shortname, a mobility-driven, context-aware collaborative learning approach that leverages user mobility to transport and update models among fixed devices, eliminating the need for stable or centralized network connections. 

    \item We validate \shortname on two distinct tasks—image classification (CIFAR-100~\cite{krizhevsky2009learning}), and human activity recognition (EgoExo4D~\cite{grauman2024ego}), demonstrating that the framework can handle diverse data distributions and modalities under limited or intermittent connectivity. We have chosen these datasets because they are widely used and therefore good benchmarks for performance and because the tasks are relevant to physical location---different locations naturally imply different subsets of features in images taken in those locations, while different activities are naturally tied to the locations in which they are more likely to be performed.
    
    \item We show that \shortname consistently outperforms
    or matches existing methods, including FedAvg~\cite{mcmahan2017communication}, Clustered Federated Learning (CFL)~\cite{sattler2019clusteredfederatedlearningmodelagnostic}, FedAS~\cite{Yang_2024_CVPR}, Gossip Learning~\cite{hegedHus2019gossip}, OppCL~\cite{9439130} and Local Only learning, across different data distributions (Dirichlet~\cite{hsu2019measuringeffectsnonidenticaldata} and Shards) under diverse mobility patterns.

\end{enumerate}

This paper is organized as follows: Section 2 reviews related work and highlights key challenges. Section 3 details the system design and underlying architecture of \shortname. Section 4 presents the evaluation methodology and results, including comparisons with existing methods. Finally, Section 5 concludes the paper by discussing the limitations of \shortname and outlining potential directions for future research.

%% file: sections/2_related.tex
During the last few years, the research community has been improving the behavior of smart enviroments to automatically adapt to user’s needs and preferences. Concretely, relevant work has two related components: (1)~distributed learning; (2)‌‌‌‌~intelligent context-aware  environments.

\textbf{Distributed learning} attempts to address the disadvantages of traditional machine learning and can be separated into two broad subcategories: federated learning (FL) and fully decentralized learning. Compared to traditional machine learning that shares data between a server and end users, federated learning keeps the data with the user. The user trains the model locally and shares updates to the model with the server~\cite{hard2018federated}. The server can then aggregate model updates from multiple users' devices using various aggregation methods, such as FedAvg~\cite{hard2018federated}, FedAS~\cite{Yang_2024_CVPR}, CFL~\cite{sattler2019clusteredfederatedlearningmodelagnostic}, etc. In contrast, fully decentralized learning functions without a central aggregation server and evolves a model with each encounter between any two mobile devices. An example in this category is Gossip Learning~\cite{hegedHus2019gossip}, which conducts an exchange-aggregate-training cycle at every encounter, while Opportunistic Collaborative Learning~\cite{9439130} conducts an exchange-training-exchange-aggregate cycle on every encounter.

Different from traditional FL, which aims to train a single global model collaboratively across multiple clients, Personalized Federated Learning (pFL) attempts to provide a personalized model or a base model that individuals can easily adapt to with few-shot learning~\cite{fallah2020personalized}. Common methods to address this include model decoupling, which splits the model structure into a shared global part (e.g., a feature extractor or backbone) and a personalized local part (e.g., task-specific classifier layers)~\cite{yi2023pfedes}, or using two models—one shared and one that remains purely local~\cite{hanzely2020federated}; meta-learning approaches that focus on finding an initial shared model that can be easily adapted to users’ local datasets with one or a few gradient descent steps~\cite{NEURIPS2020_24389bfe, Lim_2024_WACV}; adjusting the regularization function at the aggregation steps~\cite{9860349}; or using a low-rank decomposition to decouple general knowledge (shared among clients) and client-specific knowledge~\cite{10.1145/3664647.3681588}.

However, existing approaches to FL and pFL require a centralized aggregation server to coordinate aggregation, which can face single-point-of-failure issues and has high infrastructure costs in terms of computation and communication. In addition, it assumes constant internet connectivity, which is often impractical in real-world settings, where devices may only have intermittent internet access. On top of these federated learning challenges, personalized federated learning research is theoretical and focused on aggregation method and model selected, failing to utilize the dynamics of the end user and spatiotemporal context to improve model performance. Fully decentralized approaches eliminate the need for centralized aggregation but heavily rely on opportunistic and often unpredictable communication patterns, limiting their feasibility. Moreover, decentralized approaches frequently suffer from slow convergence rates due to sparse and intermittent device encounters, as well as challenges posed by heterogeneity in device environments and user behaviors. In addition,  existing decentralized learning research remains largely theoretical and has not seen widespread real-world deployment due to communication constraints~\cite{9767493}.

\textbf{Intelligent context-aware environments} are systems that sense, interpret, and respond to contextual information (e.g., user preferences, environmental conditions, temporal factors) to deliver personalized, adaptive services. The machine learning method used in such an environment is called Context-Aware Machine Learning. This approach highlights the diverse nature of contextual information and its relevance across spatio-temporal domains~\cite{10.1007/978-3-031-73110-5_17}. Harries et al. claim that one of the most important aspects of Context-Aware Machine Learning is understanding hidden properties that change over time. In the real world, a good machine learning model should not only perform good classification with the available data, but also detect changes in the hidden properties and update the model accordingly~\cite{harries1998extracting}. Sarker et al. categorize context into external or physical context, including location, time, light, movement, etc., and internal or logical context, such as a user’s interaction, goal, and social activity~\cite{Sarker2021}. Such machine learning methods are useful across different domains, such as computer vision~\cite{WANG2023103646}, human activity recognition~\cite{miranda2022survey}, smart home control~\cite{9097597}, autonomous and smart transportation~\cite{9931527}, etc.~\cite{nascimento2018context, sim2018online, liu2017context}.

Modular Machine Learning methods are commonly used in Context-Aware Learning~\cite{9931527}. They break the problem into components, solve these with different models, and combine the results into a final solution~\cite{menik2023modularmachinelearningsolution}. This breakdown method can be split into data modularity, which modularizes input for deep learning models; task modularity, which breaks down the task into sub-tasks and develops sub-models to address each problem individually; and model modularity, which focuses on the machine learning model itself~\cite{SARKER2020102762}.  Model modularity can be split into hybrid, ensemble, and graph-based Machine Learning~\cite{9931527}. In the hybrid method, researchers use two deep learning architectures and merge them in the final output~\cite{BAYOUDH2024102217}. Omolaja et al. developed context-aware human activity recognition models using a hybrid approach that utilizes light conditions and environmental noise levels with IMU signal data to predict the activity the user is performing~\cite{app12189305}. Ensemble methods are similar to hybrid methods, but the only difference is that the hybrid method directly combines the network, whereas the ensemble method post-processes multiple independent model outputs~\cite{SARKER2020102762}. Bejani et al. proposed a 
driving style evaluation system, called CADSE, which uses an ensemble method with smartphone sensors and context data. In their proposed system, the result is an ensemble of car, traffic, and maneuver classification modules~\cite{BEJANI2018303}. Graph-based methods is widely used in context-aware recommendation algorithms. For example, Wu et al. designed a graph-based multi-context-aware recommendation algorithm that uses neighborhood aggregation based on an attention mechanism with local context to enhance the representations of users and items~\cite{WU2022179}.
 
Mobile phones are among the most important devices for context-aware machine learning because they record detailed individual contextual data, including where, when, and with whom users interact during their daily activities~\cite{Sarker2021, SARKER2020102762}, and they also have sufficient computational power for deep model training and inference~\cite{9767493}. However, existing efforts do not effectively use such mobile information, often treating context-awareness as an additional weight in the model, which is ineffective. For example, a model used in a desert does not need information on how to handle rain. In addition, Brdiczka claims that a successful relationship between humans and AI requires meeting users’ expectations in diverse environments and maintaining full control over AI systems~\cite{Brdiczka_2019}. The former aligns with context-aware machine learning, which adapts to varying contextual information to provide personalized services. The latter matches the principles of distributed and collaborative learning, where data remains on the user’s device, granting users maximum control over their models and enhancing privacy.

While federated and decentralized learning address some drawbacks of traditional machine learning, they still face practical barriers such as heavy reliance on stable network  infrastructure or slow convergence in opportunistic settings. Simultaneously, context-aware machine learning underscores the importance of adapting to users’ diverse real-world conditions but often treats context as a mere additional feature rather than an integral part of model design or data distribution. Thus, existing solutions fall short of leveraging the spatial mobility of users and device interactions in a way that marries both decentralized and context-aware principles.

%% file: sections/3_system.tex
To address the challenges listed above, we propose \shortname, a mobility-driven context-aware collaborative learning approach that utilizes individual users and their mobile devices as \textit{Mules} to carry models between different physical spaces, exchanging and evolving the models alongside fixed devices located in those spaces.

In the \shortname architecture, two categories of devices are involved: mobile devices ($m_a$), which accompany users who can move between spaces, thereby acting as mules that carry the model across different locations; and fixed devices ($f_x$), which are stationary devices deployed in specific rooms or spaces. We denote the set of all mules as $M$ the set of all fixed devices as $F$.

The system provides three main functionalities: 
\begin{enumerate}
    \item \textbf{Mule}: \sloppy A mobile device $m_a$ that carries the model $w$ from $f_x$ to $f_y$ begin at time $t_j$. We donate this operation by ${\it mule}(w^{t_j}_a,m_a,f_x,f_y)$.
    \item \textbf{Hosting}: fixed device $f_x$, stores the model $w$ and allows other devices to access it (denoted by ${\it host}(w, f_x)$)
    \item \textbf{Local Training}: Depending on the data location, the training can be performed either on a mobile device $m_a$ (donated by ${\it train}_{m_a}(w)$) or on the fixed device $f_x$ (donated by ${\it train}_{f_x}(w)$).
\end{enumerate}

\begin{figure}[!t]
    \centering
    \begin{subfigure}[b]{0.48\textwidth}
        \centering
        \includegraphics[width=\textwidth]{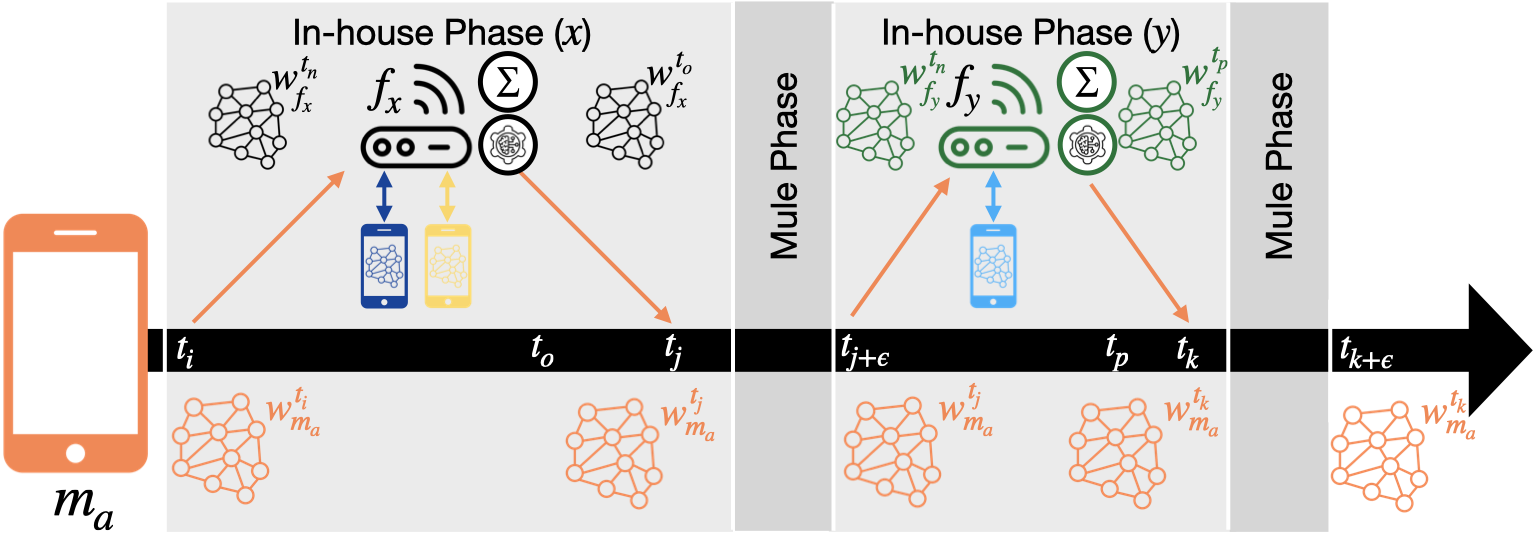}
        \caption{$f_x \in F$ and $f_y \in F$ training scenario. New data is collected on the fixed devices $F$. The mobile device $m_a \in M$ shares the model with $f_x \in F$, receives updates, mules to $f_y \in F$, and repeats the process. Each $f_x \in F$ and $f_y \in F$ both aggregate and train the model upon receiving it from $m_a \in M$.}
        \Description{Fully described in the text}
        \label{fig:in_house}
    \end{subfigure}
    \begin{subfigure}[b]{0.48\textwidth}
        \centering
        \includegraphics[width=\textwidth]{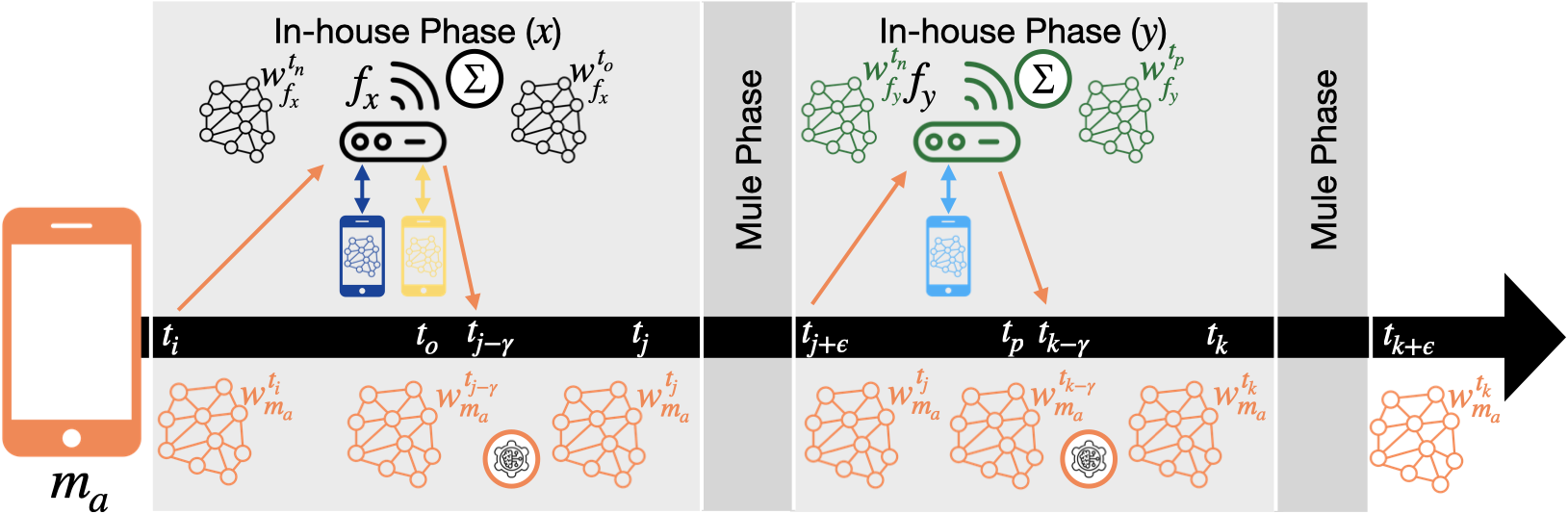}
        \caption{$m_a \in M$ training scenario. New data is collected on the mobile device $m_a$. The device shares the model with $f_x \in F$, receives the aggregated version, performs on-device training with its local data, then mules the updated model to $f_y \in F$ and repeats. In this mode, $f_x \in F$ and $f_y \in F$ only aggregate the model rather than training.}
        \Description{Fully described in the text}
        \label{fig:ondevice}
    \end{subfigure}
\vspace{-2em}
    \caption{Illustration of the two main training modes. In (a), the main training occurs on fixed devices $F$; in (b), training takes place on the mobile devices $M$.}
    \Description{Fully described in the text}
    \label{fig:system_diagram}
\vspace{-1.5em}
\end{figure}

Figure~\ref{fig:system_diagram} illustrates the overall architecture of the system. Subfigures~\ref{fig:in_house} and \ref{fig:ondevice} show two major phases of the proposed system, the \textit{in-house} phase and the \textit{mule} phase, which will be explained in more detail later in this section. Both figures are drawn with one mobile device $m_a$ at their center: the mobile device $m_a$ enters the space of the fixed device $f_x$, evolves the model with $f_x$, leaves the space, and carries the model $w^{t_j}_a$
to $f_y$ (${\it mule}(w^{t_j}_a,m_a,f_x,f_y)$), then evolves the model with $f_y$, and continues this procedure. At the same time, other devices may be interacting with $f_x$ and $f_y$ as well.

The major difference between subfigures~\ref{fig:in_house} and~\ref{fig:ondevice} is that in subfigure~\ref{fig:in_house}, the model training is completed on fixed devices $f_x$ (${\it train}_{f_x}(w)$) and $f_y$ (${\it train}_{f_y}(w)$), while in subfigure~\ref{fig:ondevice}, the model is trained on the mobile device $m_a$ (${\it train}_{m_a}(w)$). This difference arises because  training is intended to happen on the device that collects the data, but the data location depends on the downstream task. For example, if the downstream task is human activity recognition, the mobile devices $M$ conduct the training since the data is generated on those devices. Conversely, if the downstream task is smart home control, the fixed devices $F$ may have more information about the user's actions, so they conduct model training.

As shown in Figure~\ref{fig:system_diagram}, \shortname involves several interconnected steps. At the beginning, each mobile device $(m_a \in M)$ continuously detects any fixed device $(f_x \in F)$ through short-range communication protocols such as Bluetooth, Wi-Fi Direct, or other network discovery mechanisms~\cite{10298476}, and vice versa (i.e., ${\it Discover}(m_a)$ and ${\it Discover}(f_x)$). When both devices discover each other at time $t$, we consider $m_a$ and $f_x$ to be co-located, represented as $c = \langle m_a, f_x, t \rangle$. The set of all co-location events is $C = \{c\}$, and we use $C[m_a, t_i, t_j]$ to denote all co-location events involving the mule $m_a$ that occur in the time window $[t_i, t_j]$, $C[f_x, t_i, t_j]$ to denote all co-location events involving the fixed device $f_x$ in the same time window, and $C[f_x, m_a]$ to denote all co-location events involving the mule $m_a$ and fixed device $f_x$.

It is important to note that the Fixed Device Training and Mobile Device Training settings are not mutually exclusive. If needed and the application setting is appropriate, both types of training can occur in succession.

\subsection{In-House Phase}

\shortname's \textbf{In-House Phase} begins when devices $m_a$ and $f_x$ are co-located.
Specifically, if $\exists (c = \langle m_a, f_x, t_i \rangle)$ such that $\not\exists(c = \langle m_a, f_x, t_{i-1} \rangle)$, \shortname identifies time $t_i$ as the point of initial contact between the mule $m_a$ and the fixed device $f_x$. At this point in time, \shortname kicks off one of two versions of its training process.

{\bf Fixed Device Training}. For applications where fixed devices collect data and perform local training, the discovery event initiates a \emph{share-aggregate-train-share} cycle. In this process, the devices complete the following steps, in order:
\begin{enumerate}
    \item $m_a$ sends its local model weights to $f_x$ (denoted as ${\it send}(m_a, f_x, w)$)
    \item $f_x$ filters the model based on its freshness; details of this process are explained below.
    \item $f_x$ aggregates the received model with its own model (see the discussion of the aggregation process, below)
    \item $f_x$ performs ${\it train}_{f_x}(w)$ using $f_x$'s local data
    \item $f_x$ sends the updated model weights back to the mobile device $m_a$ (${\it send}(f_x, m_a, w)$) 
    \item $m_a$ aggregates the received model with its own (see the discussion of the aggregation process, below)
\end{enumerate}

{\bf Mobile Device Training.}
Alternatively, the discovery event may trigger other applications to perform training on the mobile device (i.e., the mule). In this case, the devices perform a \emph{share-aggregate-share-train} cycle, completing the following steps:
\begin{enumerate}
    \item $m_a$ sends its local model weights to $f_x$ (${\it send}(m_a, f_x, w)$)
    \item $f_x$ filters the model based on its freshness; details of this process are explained below.
    \item $f_x$ aggregates the received model with its own model (see the discussion of the aggregation process, below)
    \item $f_x$ sends the aggregated model weights back to $m_a$ (${\it send}(f_x, m_a, w)$)
    \item $m_a$ aggregates the received weights with its own
    \item $m_a$ trains the aggregated model using its local data (${\it train}_{m_a}(w)$)
\end{enumerate}
The first three steps in this second process are the same as in the Fixed Device Training approach; they serve to ensure that the mule leaves a record of having visited the space so that other mules that arrive can learn from this mule's prior experiences. This is how \shortname achieves coupling in space while also achieving {\em decoupling} in time.

\textbf{Model Freshness.} \shortname employs a filter mechanism to prevent outdated models carried by a mule from contaminating subsequent updates. This mechanism filters models using a dynamic threshold that reflects the age of a model, as measured by its last update time. Rather than using a fixed threshold, the dynamic threshold is updated as follows:
\begin{equation*}
    T^{f^{t_{i+1}}_x} = (1 - \alpha) \, T^{f^{t_{i}}_x} + \alpha \, \left(\tilde{L}^{f^{t_{i}}_x} + \beta \, \text{median}\left(\left|L^{f^{t_{i}}_x}_i - \tilde{L}^{f^{t_{i}}_x}\right|\right)\right)
\end{equation*}

Here, $\tilde{L}^{f^{t_{i}}_x} = \text{median}(L^{f^{t_{i}}_x})$, where $L^{f^{t_{i}}_x}$ is the list of model update times for device $f_x$ at time $t_i$, and $T^{f^{t_{i}}_x}$ represents the corresponding threshold. The parameters $\alpha$ and $\beta$ control the influence of the previous threshold and the variability in update times (captured by the median absolute deviation), respectively. This adaptive approach allows the system to respond to fluctuations in the environment, ensuring that only relatively fresh models are incorporated during aggregation.

{\bf Model Aggregation.} During each co-location of a mobile device $m_a$ with a fixed device $f_x$, multiple training cycles may occur. Since neither device knows when the current co-location will end, they continuously execute the following sequence: 1) the device training pipeline as described in the previous paragraphs; 2) wait for a constant delay $d$ (i.e., a user-defined or dynamically adjustable parameter); and 3) repeat for as long as $m_a$ remains co-located with $f_x$. As a result, dwell time directly influences the final aggregation weights, as devices that stay longer in a space will engage in more training cycles and more frequent parameter exchanges, thereby having a greater impact on the model evolution. In this work, we use a weighted averaging method~\cite{pmlr-v54-mcmahan17a} to aggregate the models, as such methods have been widely adopted by the research community. However, this aggregation can be easily replaced with other aggregation methods, such as FedDyn~\cite{acar2021federated}, SCAFFOLD~\cite{pmlr-v119-karimireddy20a}, or FedProx~\cite{li2020federatedoptimizationheterogeneousnetworks}. Other styles of aggregation could also consider incorporating the amount of data or the quality of the model, as presented in~\cite{yu2023idml}.

\subsection{Mule Phase}

\shortname's \textbf{Mule Phase} is much simpler. It begins when the mobile device $m_a$ physically leaves the previous environment, which is defined as $m_a$ no longer detecting $f_x$ via short-range communication protocols. During the Mule Phase, $m_a$ holds the most updated model it received in the previous In-house Phase and continues discovering other fixed devices $f_y \in F$ (${\it mule}(w, m_a, f_x, f_y)$). Once new devices are detected, the next In-house Phase begins.

Meanwhile, if there is no $m \in M$ detected by a fixed device $f_x \in F$, then $f_x$ will hold the model (${\it host}(w, f_x)$) until the next $m \in M$ is detected, triggering another In-house Phase.

%% file: sections/4_evaluation.tex
In this section, we answer the following research questions:

\begin{enumerate}
    \item Does \shortname achieve competitive convergence speed and final accuracy compared to existing distributed learning paradigms?
    \item How do different mule mobility patterns affect the overall model accuracy and convergence rate?
    \item Does the distribution of data across mobile and fixed devices influence learning outcomes and system stability?
    \item How does \shortname adapt its performance (e.g., model convergence time, final accuracy) across different downstream tasks with varying data modalities and complexities?
\end{enumerate}

\subsection{Experiment Design}
To address the research questions outlined above and evaluate the performance of \shortname, we conduct three experiments using the CIFAR-100 dataset~\cite{krizhevsky2009learning} and the EgoExo4D dataset~\cite{grauman2024ego}.  These experiments incorporate three distinct structured simulated mobility patterns and one real encounter dataset, along with prototype testing to verify the feasibility of the proposed system. The \textbf{first experiment} focuses on the fixed-device training scenario ${\it train}_{f_x}(w)$ (illustrated in Figure~\ref{fig:in_house}), aiming to answer research questions Q1 through Q3 under the setting that $f_x \in F$ conducts model training.
In contrast, the \textbf{second} and \textbf{third experiments} involve the mobile-device training scenario ${\it train}_{m_a}(w)$ (as shown in Figure~\ref{fig:ondevice}). These latter experiments utilize two distinct datasets to address \emph{all} research questions when $m_a \in M$ is responsible for training.

\begin{figure*}[t]
  \centering
  \begin{minipage}[t]{0.16\textwidth}
    \centering
    \includegraphics[width=1.2\textwidth]{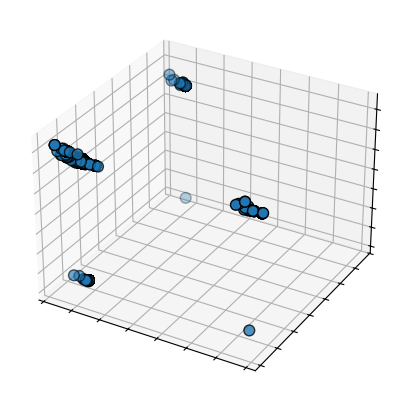}
    \vspace{-2em}
    \caption{ICA decomposition on a sample of NYC Foursquare mobility data}
    \label{fig:ICA}
  \end{minipage}
    \hspace{0.01\textwidth}
  \begin{minipage}[t]{0.8\textwidth}
    \centering
    \includegraphics[width=\textwidth]{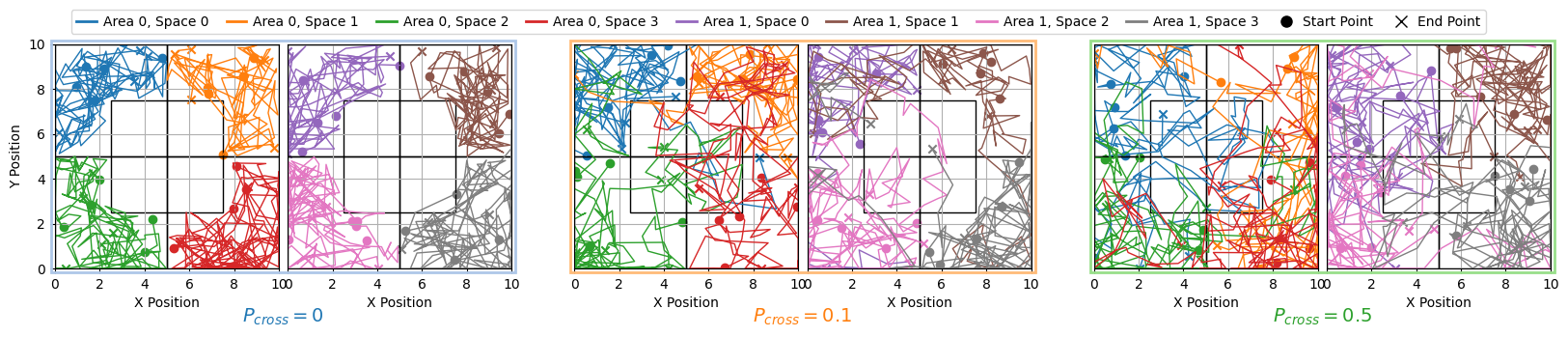}
    \vspace{-2em}
    \caption{Example random-walk trajectories under three different crossing probabilities 
      ($P_{cross} = \{0, 0.1, 0.5\}$). Each subfigure shows device movements in a 2D space 
      partitioned into four spaces within two isolated areas. The black grid lines mark boundaries, 
      while circles and crosses denote start and end points, respectively. Higher $P_{cross}$ values 
      indicate greater likelihood of leaving the current space.}
    \label{fig:random_walk}
  \end{minipage}
\vspace{-1em}
\end{figure*}

We first examined a real-world human mobility dataset collected by Foursquare \cite{Foursquare} for two cities, New York City, NY (NYC) and Austin, TX. The dataset has been de-identified and assigns a unique identifier to each person and location. It contains information on when a specific user visited a particular place and for how long, based on a subset of Foursquare app users' mobility and covering the period from 2018 to 2020. We specifically focused on the 2018 data, as it predates the COVID pandemic and the encounters were less affected by lockdowns. For one month of data, NYC includes 127,242 unique individuals, 144,274 different places, and 4,699,150 unique data points; Austin includes 20,076 unique individuals, 17,625 different places, and 505,016 unique data points.

We examined the Foursquare data and observed that most individuals consistently visit a specific subgroup of locations while rarely going to others. An independent component analysis (ICA) of the frequently visited places by various individuals, based on the NYC data, is shown in Figure~\ref{fig:ICA}. For instance, the cluster of people in the top left of Figure~\ref{fig:ICA} visit a similar set of locations and differ from the group in the bottom right. Inspired by this observation, we modeled human mobility using a random-walk approach, wherein each device moves freely within a space, making one move per time step (with "time step" serving as the basic unit for measuring all actions).
We introduce a parameter $P_{cross}$ that controls the probability of leaving the current space. This design simulates real-world human interactions in which a user or device predominantly interacts within a subset of the entire area and occasionally transitions to another region, effectively creating the `subgroup' phenomenon we observed in the Foursquare dataset.

We further observed that only a very small fraction of individuals travel between NYC and Austin (approximately $0.715\%$ of the total population in the dataset). Thus, in each simulation, we designed two square
areas that are completely isolated from one another, i.e., devices cannot communicate across the isolated areas. Within each area, we define \emph{four spaces}, and we assume there are 8 \emph{fixed devices} $f_x \in F$ (one deployed in each space). The central space of the area remains empty (i.e., it does not contain any fixed devices) and does not overleap to any of the four surrounding spaces.

Such a design mimics real-world mobility patterns, where users may: (1) directly move from one room to another (cross spaces directly); (2) leave one space, traverse an open space, and enter another space (cross spaces via central empty area); or (3) interact exclusively within certain spaces (e.g., in Area 0) without ever entering other spaces (e.g., in Area 1).

An example of the random-walk pattern, under three different values of $P_{cross}$, is presented in Figure~\ref{fig:random_walk}.
Each trajectory is colored based on the device's starting location and illustrates how devices traverse different spaces, with start points (circles) and end points (crosses) indicated in each subfigure. We assume that the fixed devices ($f_x \in F$) are installed at the center of each space and can, and only can, communicate with devices within their respective spaces via peer-to-peer methods.

The simulated patterns are designed for simplified simulation, to facilitate future research, and to understand how various factors affect model performance. In this paper, we conduct experiments using both real and simulated patterns.

\subsection{Evaluation with Fixed-Device Training}
We begin by evaluating the performance of \shortname when model training is conducted on fixed devices ($f_x \in F$). We use the CIFAR-100 dataset~\cite{krizhevsky2009learning} along with the mobility patterns described above. To provide a comprehensive comparison, we benchmark \shortname against four baselines:

\begin{itemize}
    \item FedAvg~\cite{mcmahan2017communication}: A traditional federated learning algorithm
    \item FedAS~\cite{Yang_2024_CVPR}: A recently proposed personalized federated learning approach
    \item CFL~\cite{sattler2019clusteredfederatedlearningmodelagnostic}: A clustered federated learning method
    \item Local-only: Each device trains locally without any communication
\end{itemize}

\subsubsection{Experimental Setup}
\begin{figure*}[t]
  \centering
  \includegraphics[width=0.98\textwidth]{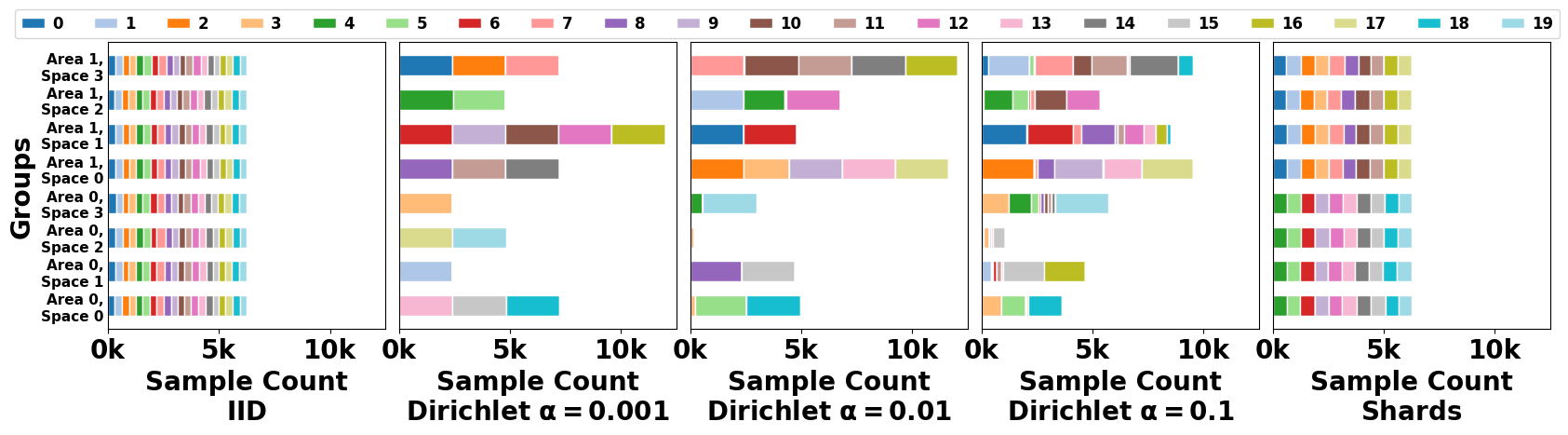}
    \vspace{-1em}
  \caption{CIFAR-100 Data distributions across different partitioning methods.
    The first subplot show IID distribution. Next three subplots illustrate Dirichlet-based distributions with $\alpha= \{0.001, 0.01, 0.1\}$.
    The last subplot shows our adapted Shards method, wherein super-classes are split between two areas (Area~0 and Area~1), and each space within an area contains exactly one subclass.
    }
    \Description{Fully described in the text}
  \label{fig:data_dist} 
    \vspace{-1em}
\end{figure*}

CIFAR-100 comprises 20 super-classes and 100 classes, with each super-class containing exactly 5 classes. In our experiments, we use these 20 super-classes as the classification targets. We distribute the dataset in 5 distinct ways, including  independent and identically distributed (i.i.d.) and non-iid as illustrated in Figure.~\ref{fig:data_dist}. The non-iid distributions employ a Dirichlet-based partitioning scheme~\cite{hsu2019measuringeffectsnonidenticaldata}, where smaller $\alpha$ values typically yield a distribution closer to iid setting.

In this experiment, we use a lightweight convolutional neural network (CNN) 
for efficient training in resource-constrained environments. The architecture features a feature extractor with two convolutional blocks (3×3 convolution, batch normalization, ReLU activation, and pooling) and a classifier with two fully connected layers.

During the simulation, all baseline federated learning methods collect models from their respective clients and subsequently perform aggregation and send the global model back to clients. We assume that model sharing is completed within one time step and define this process as one round of model evolution. For \shortname, however, due to the opportunistic nature of encounters and peer-to-peer communication, model exchange does not happen instantly and is generally slower than the internet. We assume that sharing a model takes three time steps.
Defining such a round is more challenging. In this experiment, we deploy 20 mobile devices in the environment and define one round of model evolution as 20 successful peer-to-peer model exchanges. Due to the nature of peer-to-peer communication, some devices may not enter any space during every round. The 20 peer-to-peer model exchanges might not involve every device, and some devices might communicate more than once.
Meanwhile, in the Local-only method, each device does not communicate with any other device; thus, one round of training on each devices is one round of model evolution. 

In all experiments, we use $20\%$ of the data on each fixed device as a testing 
dataset to test how the model would perform in the space; this testing data is excluded from all training steps but has the same data distribution as the training dataset. Prior to the simulation, the model is pretrained on its assigned training data
until the testing accuracy stops improving. The performance has been evaluated after the model returned to the fixed devices, and was retrained for 1 epoch with local training data as a fine-tuning step, evaluated on its testing data, and accuracy was used as the primary metric to evaluate the model's performance. We report the accuracy of the model before local training on the baseline methods to align our results with other federated learning research.

\subsubsection{Results and Discussion}

\input{sections/5_evaluation_table}

Table~\ref{tab:fixed_device_result}
summarizes the performance of our proposed approach under both IID and non-IID data distributions (modeled using a Dirichlet distribution with $\alpha \in \{0.001, 0.01, 0.1\}$) for each method—CFL~\cite{sattler2019clusteredfederatedlearningmodelagnostic}, FedAS~\cite{Yang_2024_CVPR}, FedAvg~\cite{mcmahan2017communication}, Local Only, and our \shortname. These baseline methods are not affected by mobility patterns; therefore, each yields only a single accuracy value (one before local training and another after local training). On the other hand, since \shortname relies on the mobility patterns of mobile devices, the bottom rows indicate \shortname's accuracy under three different mobility patterns, $P_{cross} \in \{0, 0.1, 0.5\}$, as well as real encounter data from the Foursquare dataset.

As we assume that each fixed device’s model is used only within its designated space, accuracy is evaluated on a dataset matching that area’s data distribution. By incorporating an additional local training round after the model returns to a fixed device, the model becomes more specialized to its local data, potentially achieving even higher accuracy than a globally optimized model. However, such a locally specialized model may not reflect the {\em global} optimum, as what is optimal globally may differ from what is optimal locally.

In these evaluations, a larger value of Dirichlet $\alpha$ implies a more non-IID distribution. Figure~\ref{fig:data_dist} shows that varying $\alpha$ also affects the number of categories each fixed device specializes in. For instance, in area 1, space 3, when $\alpha = 0.001$, the local model may only need to differentiate between three classes, whereas with $\alpha = 0.01$, it must handle five classes, and nine classes for $\alpha = 0.1$, thereby increasing task complexity and reducing model performance across all methods. In the IID scenario, the model must recognize all 20 classes, making it the most challenging setting. Unsurprisingly, it achieves the lowest accuracy compared to the non-IID conditions.

Comparing \shortname to the baseline methods, \shortname consistently outperforms or remains competitive with them. We attribute this improvement to more effective clustering of models, enabled by the movement of mobile devices (mules), which provide context-aware information to fixed devices. This context-awareness allows fixed devices to gain a deeper understanding of their specific objectives, ultimately leading to better model performance.

Additionally, we observe that $P_{cross}$ influences ML-Mule’s performance. A higher $P_{cross}=0.5$ facilitates more frequent inter-space movement, allowing devices to access a wider variety of models. However, it can also introduce instability in the early training stages, as mobile devices bring diverse models from other spaces into the current space. When $P_{cross}=0$, devices never leave their designated area, meaning that the data each device can access is limited to local training data. Despite this, the model still performs better than the Local Only setting, likely due to the different-stage weight averaging effect. Some mobile devices have a slightly earlier stage of the model ($w^{t_{i-3}}$), which is shared back to the fixed device at $t_i$, similar to what was observed in previous work~\cite{izmailov2018averaging}.

The simulation using real Foursquare dataset encounters is presented in the last column of simulation results, indicated by the column labeled `4Q'. A similar or slightly lower performance is observed with the Foursquare dataset compared to the simulated patterns. This is primarily because the simulated patterns were designed to be denser to reduce computational requirements and demonstrate the general performance of the proposed approach. In contrast, the raw Foursquare data is more sparse—many mules appear briefly and then disappear, without sustained participation. However, these results demonstrate the feasibility of the proposed method and confirm that the simulated patterns reasonably reflect real-world performance. For the simulated mobility pattern, we conducted experiments with various random seeds and obtained similar results.

Unlike other federated learning methods that depend on cloud connectivity and existing network infrastructure, \shortname evolves models exclusively through local, peer-to-peer exchanges. This approach provides a significant advantage in real-world environments with intermittent or nonexistent internet access. \shortname consistently outperforms the Local Only method, which, aside from \shortname, may be the only viable option when stable or high-speed internet is unavailable. These advantages are particularly valuable in environments with limited or even non-existent connectivity, simplifying the deployment and setup of devices. For instance, a small, low-cost device (e.g., a Raspberry Pi) could be deployed without internet setup or additional configuration. Using \shortname, the model would still evolve through opportunistic encounters with mobile devices, ensuring continuous improvement even in disconnected settings.

\subsection{Evaluation with Mobile-Device Training}
This section examines how \shortname handles model training when new data primarily resides on mobile devices ($m_a \in M$) that frequently move across different spaces. In these experiments, as shown in Figure~\ref{fig:ondevice}, the fixed devices ($f_x \in F$) act solely as model holders, receiving, aggregating, and returning the model without performing any local training. Conversely, the mobile devices ($m_a \in M$) exchange the model with these fixed devices and train it on their own local data.

As this setting is more aligned with fully decentralized learning than classical federated learning, we compare \shortname with Gossip Learning~\cite{hegedHus2019gossip} and OppCL~\cite{9439130}, two decentralized learning approaches, and the Local-only method. We again use accuracy as our performance metric. We also evaluate a combination of the proposed approach with Gossip Learning, as the two approaches can operate orthogonally. 

\subsubsection{Experimental Setup}

We first use CIFAR-100, focusing on the 20 super-classes as classification targets. Data is allocated to mules rather than fixed devices, following the Shards approach used in the original FedAvg paper~\cite{mcmahan2017communication}, as shown in Figure~\ref{fig:data_dist}. To better reflect real-world conditions, we further split the subclasses across different spaces. First, we evenly divide the 20 super-classes between Area~0 and Area~1, ensuring no overlap. Within each area, we then assign exactly 1 subclass of a given super-class to each of the four spaces (again with no overlap). For instance, if Area~0 contains the ``vehicles~1'' super-class, then: Space~0 in Area~0 might contain only the ``bicycle'' subclass; Space~1 might contain only ``bus'' subclass; etc. Since each super-class has 5 subclasses, and we define only four spaces, the fifth subclass is omitted in this setup. 

Depending on a device’s initial space, it receives 2500 images from that space’s distribution (representing specialized local information) and an additional 2500 images from the fifth class in the assigned super-class (representing more general knowledge). To accommodate resource-constrained devices, we employ the same lightweight CNN model from the previous experiment. In addition, at every time step, each mobile device acquires a new image from its current space, reflecting ongoing data generation in realistic scenarios.

As a second dataset, we use EgoExo4D~\cite{grauman2024ego}, which is a multi-modal, multi-view video dataset recorded across 13 global sites. We extract and focus on the IMU data (accelerometer and gyroscope readings), commonly used in HAR tasks. The sensor data is down-sampled at 50Hz. To handle sequential IMU data, we employ an LSTM-CNN model structure, which is well-established in HAR research~\cite{9043535}.

To reduce simulation complexity, we focus on several distinct indoor activities from the top eight locations in EgoExo4D. Table~\ref{tab:IMU_Data_Dist} summarizes the distribution of data across spaces. Each simulation space is randomly assigned a corresponding location, and the data associated with that location is provided to the simulation space. The numbers in the table represent the number of separate data collection sessions conducted at each location for the specified activity classes. Unlike the image classification scenario, no additional data is introduced during the simulation; models evolve solely through interactions between mobile and fixed devices and retraining with the data they already have.

\begin{table}[t]
\caption{Distribution of IMU data point across different locations for various activities.}
\vspace{-1em}
\label{tab:IMU_Data_Dist}
\resizebox{0.475\textwidth}{!}{
\begin{tabular}{|c|c|c|c|c|c|c|c|c|}
\hline
\textbf{Class/Location} & \textbf{cmu} & \textbf{fair} & \textbf{gt} & \textbf{iiith} & \textbf{indiana} & \textbf{sfu} & \textbf{uniandes} & \textbf{upenn} \\ \hline
\textbf{Bike Repair}     & 72           & 71            & 109                  & 0              & 107              & 0            & 0                 & 0              \\ \hline
\textbf{Cooking}        & 0            & 17            & 64                   & 222            & 64               & 98           & 44                & 69             \\ \hline
\textbf{Dance}          & 0            & 0             & 0                    & 0              & 0                & 0            & 576               & 152            \\ \hline
\textbf{Music}          & 0            & 0             & 0                    & 16             & 49               & 0            & 0                 & 203            \\ \hline
\end{tabular}}
\vspace{-1em}
\end{table}

At each simulation time step, Gossip Learning, OppCL, and the Gossip component of ML Mule+Gossip allow mobile devices to attempt communication with surrounding mobile devices within a defined communication radius, sharing their models. We assume that it takes 3 time steps
to completely share a model with neighboring devices via the peer-to-peer network. Similarly, in \shortname and the Mule component of the \shortname+ Gossip method, mobile devices require 3 time steps to share their models with the fixed devices. These communications are limited to the fixed device assigned to the space where the mobile device is currently located. In the Local Only method, each mobile device trains its model with its own training data for one epoch at each time slot. 

In both simulations, we hold out 20\% of the data for testing. Prior to the simulation, devices pretrain the model on their local data until no further improvement is observed on their local test sets. Each mule moves through the environment according to the random-walk model introduced earlier. The evaluation at any point in time is based on the data from the space where a device is currently located.

\subsubsection{Results and Discussion}

\begin{figure*}[t]
  \centering
  \includegraphics[width=0.98\textwidth]{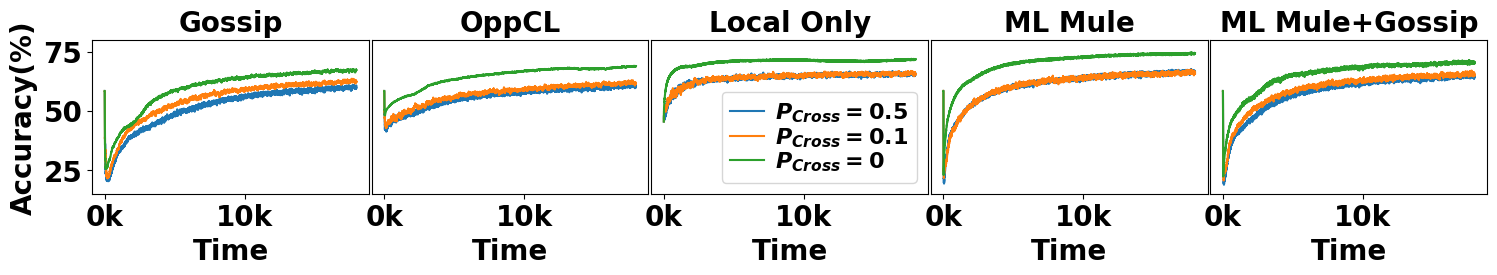}
   \vspace{-\baselineskip}
  \caption{Accuracy over time for image classification with different methods and crossing probabilities. A higher $P_{cross}$ indicates more frequent movement between spaces, while $P_{cross} = 0$ implies no cross.
  }
  \label{fig:IMG_Result} 
  \Description{Fully described in the text}
\end{figure*}

\begin{figure*}[t]
  \centering
  \includegraphics[width=0.98\textwidth]{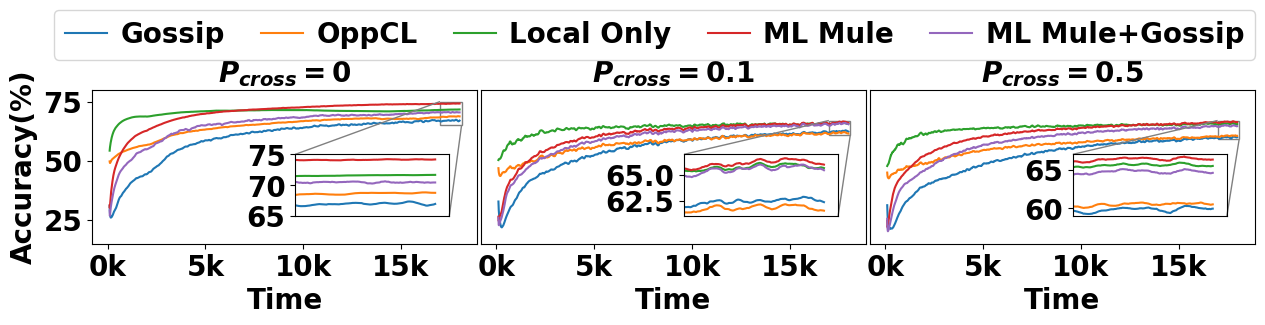}
   \vspace{-\baselineskip}

  \caption{Accuracy over time for image classification with different crossing probabilities and methods. To reduce the overlap between lines, a moving average with 100 time steps was applied to minimize the noise in the results.}
  \label{fig:IMG_Result_p_cross} 
  \Description{Fully described in the text}
\end{figure*}

\begin{figure*}[t]
  \centering
  \includegraphics[width=0.98\textwidth]{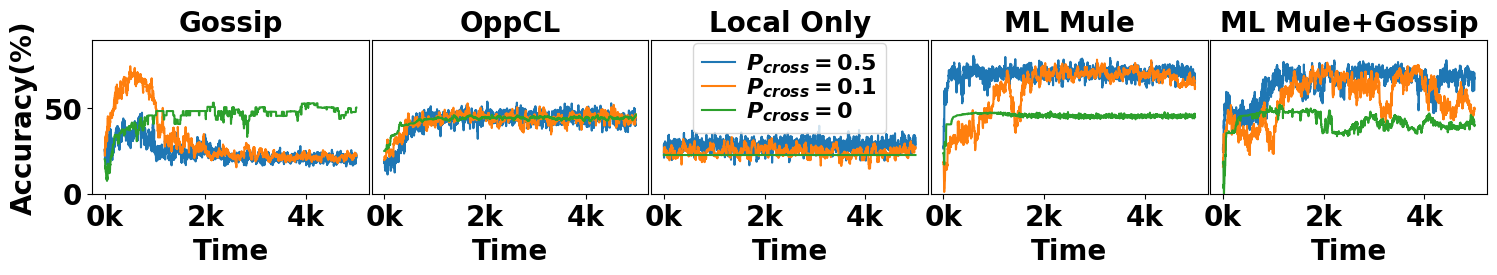}
   \vspace{-\baselineskip}
  \caption{Accuracy over time for human activity recognition with different methods and crossing probabilities. A higher $P_{cross}$ indicates more frequent movement between spaces, while $P_{cross} = 0$ implies no cross.}
  \label{fig:IMU_Result} 
  \Description{Fully described in the text}
\end{figure*}

\begin{figure*}[t]
\vspace{-.25cm}
  \centering
  \includegraphics[width=0.98\textwidth]{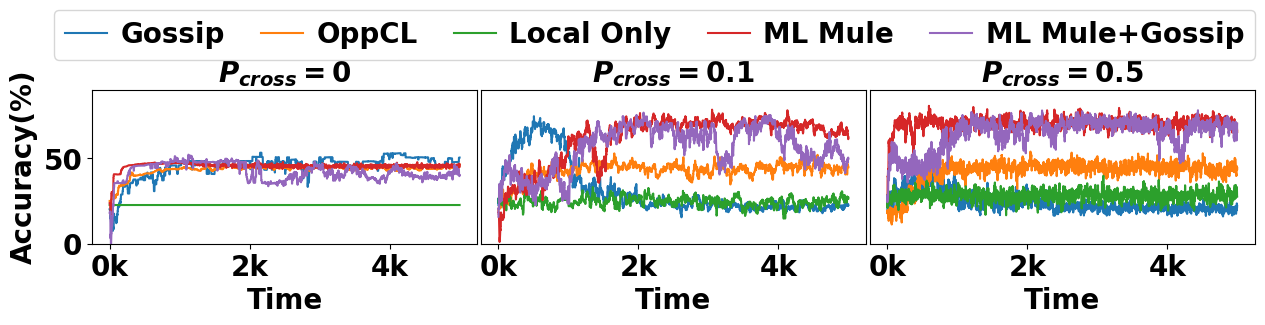}
   \vspace{-\baselineskip}

  \caption{Accuracy over time for human activity recognition with different crossing probabilities and methods.}
  
  \label{fig:IMU_Result_p_cross}
  \Description{Fully described in the text}
\end{figure*}

Figures~\ref{fig:IMG_Result} and~\ref{fig:IMU_Result} display the accuracy over time for four learning methods—Gossip, OppCL, Local Only, \shortname, and \shortname + Gossip—under three mobility patterns characterized by $P_{cross} = \{0.5, 0.1, 0\}$, using the CIFAR100 and EgoExo4D datasets. Each subfigure corresponds to one method, with the $x$-axis denoting simulation time and the $y$-axis showing the test accuracy. 

To facilitate comparison from another perspective, Figures~\ref{fig:IMG_Result_p_cross} and~\ref{fig:IMU_Result_p_cross}
reorganize the same data by plotting accuracy for different $P_{cross}$ values, where each subfigure contains four curves corresponding to the four methods. 

A consistent finding is that \shortname achieves higher accuracy in most configurations, converging more rapidly and reaching better final accuracy than other baselines. The only exception appears in the very early stages, with image classification task, where Local Only can temporarily outperform \shortname. This advantage quickly disappears once the decentralized methods stabilize their aggregated models. Such advantage does not exist in the results for the HAR task, as the task is more complex and it is difficult for the Local Only method to extract sufficient features from the limited data.

Mobility heavily influences the learning dynamics in this setting as well. When $P_{cross}=0$, devices remain in their initially assigned spaces, resulting in faster local convergence but reduced exposure to diverse data. In contrast, higher crossing probabilities such as $P_{cross}=0.5$ provide more inter-space travel, yielding a richer variety of model but sometimes causing the model accuracy fluctuations in early stage.

In the image classification task, \shortname, OppCL, and Gossip exhibit similar learning trajectories, though Gossip tends to converge more slowly and reaches lower peak accuracy. This discrepancy arises from the absence of a stable anchor (a fixed device) in pure Gossip, limiting its capacity to unify regional patterns. In \shortname, mobile devices coordinate with fixed devices that hold relatively stable local models, which capture  space-specific features. By contrast, Gossip relies on device-to-device exchanges alone, which depend on the devices it encounters and the previous experience of such devices, which can lead to slower convergence.
The HAR task reveals that Gossip Learning may struggle to obtain sufficient models from neighbors, particularly when it moves across different data distributions, resulting in performance close to that achieved by a local training-only setting. In contrast, OppCL and Local-Only are stable but do not yield improvements. ML Mule+Gossip performs similarly to \shortname, but with greater variance, suggesting that additional peer-to-peer exchanges may not significantly enhance performance once devices can already interact effectively with fixed devices.

There is no detailed movement pattern for individual users in the Foursquare data; it only records when a given user enters a space. Consequently, it is not possible to construct a realistic comparison between the decentralized learning method and \shortname. Therefore, this simulation is based on a simulated mobility pattern only. We have also tested with different random seeds and numbers of devices in the system; the results remain relatively similar unless the number of devices is insufficient to enable decentralized learning.

 \subsection{Prototype}

To evaluate the usability of our proposed system, we designed a prototype using two Jetson Orin Nano Developer Kit 8GB~\cite{jetson} and one Raspberry Pi 5 8GB~\cite{rp5}. All devices use the Wi-Fi board provided with the developer kit (for the Jetson) or the onboard Wi-Fi (for the Pi 5) as the communication hardware, and they are configured in Wi-Fi Ad Hoc mode at a frequency of 5.18 GHz (channel 36). In our setup, the Jetson Orin Nano devices serve as fixed devices in separate rooms, while the Raspberry Pi acts as the mule. For the model and training steps, we employ the same approach as in previous simulations.

The Jetson devices are placed in different spaces and cannot directly communicate with each other via the Ad Hoc Wi-Fi network, whereas the Raspberry Pi is carried by testers and powered by a battery. To ensure accurate time synchronization, the Pi’s Wi-Fi is disabled before entering a room; once inside, the Wi-Fi is enabled and time counting begins. We use the moments when the Jetson devices and the Pi detect each other as time synchronization points and compute the average duration of each action. 

\begin{figure}[!t]
    \centering
    \includegraphics[width=0.48\textwidth]{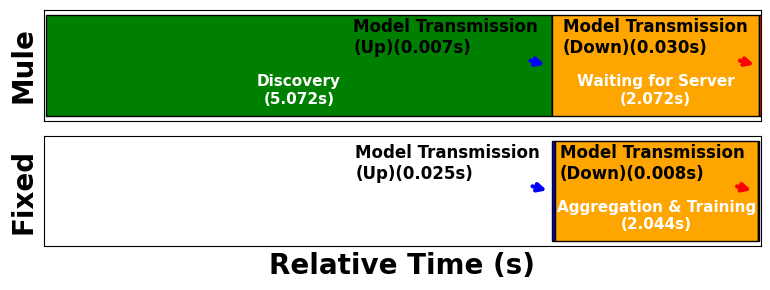}
    \vspace{-2.5em}
    \caption{Timeline of operations for the Mule device (top) and the Fixed device (bottom)}
    \Description{Fully described in the text}
    \label{fig:prototype}
    \vspace{-1em}
\end{figure}

The results are shown in Figure~\ref{fig:prototype}. The Mule spends approximately 5.07 seconds discovering the Fixed device before transmitting its model upstream, which takes 0.007 seconds, waiting for the Jetson to respond (2.07 seconds), and finally receiving the updated model back in 0.007 seconds. 

Second, we further tested our proposed system by placing all prototype devices in the same space. From the Foursquare dataset, we randomly selected a person and two places this person had visited to serve as the trajectory for the prototype experiment. The fixed devices turned their Wi-Fi on and off following the trajectory, allowing the Mule to communicate accordingly (prototyping the person’s entry into and exit from each space). We distributed data on each individual fixed device identical to what we used in previous simulations with $\alpha=0.1$.

\begin{figure}[!t]
    \centering
    \includegraphics[width=0.48\textwidth]{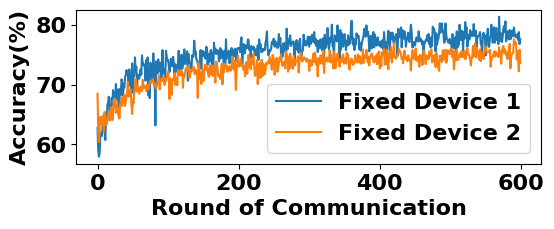}
    \vspace{-2.5em}
    \caption{Prototype \shortname with 2 fixed devices (Jetson) and 1 mule (Raspberry Pi)}
    \Description{Fully described in the text}
    \label{fig:prototype_acc}
    \vspace{-1em}
\end{figure}

The results are shown in Figure~\ref{fig:prototype_acc}. The X-axis represents the communication rounds that occur on the Mule side. Because we only have one Mule in the prototype, the communication rounds are higher; however, in real-world applications, they will be averaged across multiple Mules. We observed that the prototype system achieves performance similar to the simulated results. During each in-house phase, the CPU usage on the Mule is only around 0.7\%. These results demonstrate the feasibility of the proposed approach.

%% file: sections/5_evaluation_table.tex
\setlength{\tabcolsep}{4pt}

\begin{table*}[]
\centering
\caption{Accuracy comparison under the fixed-device training scenario. Each block represents a different distribution strategy (Dirichlet $\alpha=\{0.001, 0.01, 0.1\}$). The columns show the final accuracy after the model performance stops improving for 10 consecutive rounds. Pre-Local indicates the accuracy of the model directly received from the central server, and Post-Local refers to the accuracy after one round of local training. The top rows list baseline methods, while the bottom rows display results for \shortname at crossing probabilities $P_{cross} = \{0, 0.1, 0.5\}$, and FourSquare (4Q). All values represent the average test-set accuracy (\%), where higher is better.}
\vspace{-1em}
\label{tab:fixed_device_result}
\resizebox{\textwidth}{!}{
\begin{tabular}{|c|cccc|cccc|cccc|cccc|}
\hline
\rowcolor[HTML]{C0C0C0} 
\textbf{}                                & \multicolumn{4}{c|}{\cellcolor[HTML]{C0C0C0}\textbf{Dirichlet ($\alpha = 0.001$)}}                                                                                                                                            & \multicolumn{4}{c|}{\cellcolor[HTML]{C0C0C0}\textbf{Dirichlet ($\alpha = 0.01$)}}                                                                                                                                                  & \multicolumn{4}{c|}{\cellcolor[HTML]{C0C0C0}\textbf{Dirichlet ($\alpha = 0.1$)}}                                                                                                                                                   & \multicolumn{4}{c|}{\cellcolor[HTML]{C0C0C0}\textbf{IID}}                                                                                                                                                                                      \\ \hline
\rowcolor[HTML]{C0C0C0} 
\textbf{Round}                           & \multicolumn{2}{c|}{\cellcolor[HTML]{C0C0C0}\textbf{Pre-Local}}                                                                                       & \multicolumn{2}{c|}{\cellcolor[HTML]{C0C0C0}\textbf{Post-Local}}                                            & \multicolumn{2}{c|}{\cellcolor[HTML]{C0C0C0}\textbf{Pre-Local}}                                                                                       & \multicolumn{2}{c|}{\cellcolor[HTML]{C0C0C0}\textbf{Post-Local}}                                            & \multicolumn{2}{c|}{\cellcolor[HTML]{C0C0C0}\textbf{Pre-Local}}                                                                                       & \multicolumn{2}{c|}{\cellcolor[HTML]{C0C0C0}\textbf{Post-Local}}                                            & \multicolumn{2}{c|}{\cellcolor[HTML]{C0C0C0}\textbf{Pre-Local}}                                                                                       & \multicolumn{2}{c|}{\cellcolor[HTML]{C0C0C0}\textbf{Post-Local}} \\ \hline
CFL~\cite{sattler2019clusteredfederatedlearningmodelagnostic}                                       & \multicolumn{2}{c|}{33.19}                                                                                                                                        & \multicolumn{2}{c|}{84.58}                                                                                            & \multicolumn{2}{c|}{39.62}                                                                                                                                        & \multicolumn{2}{c|}{85.13}                                                                                            & \multicolumn{2}{c|}{47.59}                                                                                                                                        & \multicolumn{2}{c|}{74.82}                                                                                            & \multicolumn{2}{c|}{40.47}                                                                                                                                        & \multicolumn{2}{c|}{38.87}                                                 \\ \hline
FedAS~\cite{Yang_2024_CVPR}                                    & \multicolumn{2}{c|}{35.00}                                                                                                                                        & \multicolumn{2}{c|}{84.00}                                                                                            & \multicolumn{2}{c|}{38.20}                                                                                                                                        & \multicolumn{2}{c|}{83.68}                                                                                            & \multicolumn{2}{c|}{44.35}                                                                                                                                        & \multicolumn{2}{c|}{74.05}                                                                                            & \multicolumn{2}{c|}{41.01}                                                                                                                                        & \multicolumn{2}{c|}{39.47}                                                 \\ \hline
FedAvg~\cite{mcmahan2017communication}                                    & \multicolumn{2}{c|}{32.45}                                                                                                                                        & \multicolumn{2}{c|}{84.50}                                                                                            & \multicolumn{2}{c|}{36.10}                                                                                                                                        & \multicolumn{2}{c|}{84.49}                                                                                            & \multicolumn{2}{c|}{46.83}                                                                                                                                        & \multicolumn{2}{c|}{74.48}                                                                                            & \multicolumn{2}{c|}{39.97}                                                                                                                                        & \multicolumn{2}{c|}{37.57}                                                 \\ \hline
Local Only                                    & \multicolumn{4}{c|}{84.47}                                                                                                                                                                                                                                                                & \multicolumn{4}{c|}{84.70}                                                                                                                                                                                                                                                                & \multicolumn{4}{c|}{71.62}                                                                                                                                                                                                                                                                & \multicolumn{4}{c|}{35.13}                                                                                                                                                                                                                     \\ \hline
\rowcolor[HTML]{C0C0C0} 
{ $\mathbf{P_{cross}}$} & \multicolumn{1}{c|}{\cellcolor[HTML]{C0C0C0}{ \textbf{0}}} & \multicolumn{1}{c|}{\cellcolor[HTML]{C0C0C0}{ \textbf{0.1}}} & \multicolumn{1}{c|}{\cellcolor[HTML]{C0C0C0}{ \textbf{0.5}}} & { \textbf{4Q}} & \multicolumn{1}{c|}{\cellcolor[HTML]{C0C0C0}{ \textbf{0}}} & \multicolumn{1}{c|}{\cellcolor[HTML]{C0C0C0}{ \textbf{0.1}}} & \multicolumn{1}{c|}{\cellcolor[HTML]{C0C0C0}{ \textbf{0.5}}} & { \textbf{4Q}} & \multicolumn{1}{c|}{\cellcolor[HTML]{C0C0C0}{ \textbf{0}}} & \multicolumn{1}{c|}{\cellcolor[HTML]{C0C0C0}{ \textbf{0.1}}} & \multicolumn{1}{c|}{\cellcolor[HTML]{C0C0C0}{ \textbf{0.5}}} & { \textbf{4Q}} & \multicolumn{1}{c|}{\cellcolor[HTML]{C0C0C0}{ \textbf{0}}} & \multicolumn{1}{c|}{\cellcolor[HTML]{C0C0C0}{ \textbf{0.1}}} & \multicolumn{1}{c|}{\cellcolor[HTML]{C0C0C0}\textbf{0.5}} & \textbf{4Q}    \\ \hline
\textbf{ML Mule}                   & \multicolumn{1}{c|}{\textbf{91.18}}                                            & \multicolumn{1}{c|}{\textbf{89.79}}                                              & \multicolumn{1}{c|}{\textbf{89.16}}                                              & \textbf{88.95}                     & \multicolumn{1}{c|}{\textbf{91.12}}                                            & \multicolumn{1}{c|}{\textbf{89.60}}                                              & \multicolumn{1}{c|}{\textbf{88.23}}                                              & \textbf{89.05}                     & \multicolumn{1}{c|}{\textbf{75.61}}                                            & \multicolumn{1}{c|}{\textbf{76.45}}                                              & \multicolumn{1}{c|}{\textbf{75.45}}                                              & 73.89                              & \multicolumn{1}{c|}{\textbf{45.07}}                                            & \multicolumn{1}{c|}{\textbf{52.62}}                                              & \multicolumn{1}{c|}{\textbf{50.78}}                       & \textbf{48.88} \\ \hline
\end{tabular}
}
\end{table*}

%% file: sections/6_conclusion.tex
This paper introduces \shortname, a mobile-driven, context-aware collaborative learning framework that utilizes mobile devices to transport models between different spaces equipped with fixed devices, enabling decentralized model training in environments with limited or no internet connectivity. By relying on localized communication and mobile devices acting as ``mules'' \shortname enables efficient model evolution without requiring centralized infrastructure. Our evaluations on two tasks—image classification with CIFAR-100 and human activity recognition with EgoExo4D—as well as a prototype system, demonstrate that \shortname consistently outperforms or matches baseline federated and decentralized learning methods in both accuracy and convergence speed.

Despite these achievements, several limitations must be acknowledged. First, \shortname was tested using a simple weighted averaging aggregation strategy; exploring more advanced methods tailored for non-i.i.d. data distributions could further enhance its performance. Second, the experiments were conducted in relatively small-scale simulated environments. Deploying \shortname in real-world, large-scale IoT systems would require addressing additional challenges, including communication delays, hardware heterogeneity, and other resource constraints. Future researchers could expand the integration of federated learning with \shortname; incorporate privacy-preserving techniques such as differential privacy or secure multiparty computation; and conduct large-scale real-world deployments testing. 

In conclusion, \shortname demonstrates the potential of using distributed learning in a setting where coupling occurs only in space while remaining decoupled in time. Furthermore, by employing mobile devices as mules to transport the model between fixed devices, \shortname offers a robust alternative to both centralized and traditional distributed learning in resource-constrained environments.

%% file: references.bib
@inproceedings{hegedHus2019gossip,
  title={Gossip learning as a decentralized alternative to federated learning},
  author={Heged{\H{u}}s, Istv{\'a}n and Danner, G{\'a}bor and Jelasity, M{\'a}rk},
  booktitle={Proc. of DAIS},
  pages={74--90},
  year={2019},
}

@INPROCEEDINGS{9767493,
  author={Yu, Haoxiang and Chen, Hsiao-Yuan and Lee, Sangsu and Zheng, Xi and Julien, Christine},
  booktitle={Proc. of PerCom Workshops}, 
  title={Prototyping Opportunistic Learning in Resource Constrained Mobile Devices}, 
  year={2022},
  volume={},
  number={},
  pages={521-526},
}

@INPROCEEDINGS{9439130,
  author={Lee, Sangsu and Zheng, Xi and Hua, Jie and Vikalo, Haris and Julien, Christine},
  booktitle={Proc. of PerCom}, 
  title={Opportunistic Federated Learning: An Exploration of Egocentric Collaboration for Pervasive Computing Applications}, 
  year={2021},
  volume={},
  number={},
  pages={1-8},
}

@article{hard2018federated,
  title={Federated learning for mobile keyboard prediction},
  author={Hard, Andrew and Rao, Kanishka and Mathews, Rajiv and Ramaswamy, Swaroop and Beaufays, Fran{\c{c}}oise and Augenstein, Sean and Eichner, Hubert and Kiddon, Chlo{\'e} and Ramage, Daniel},
  journal={arXiv preprint arXiv:1811.03604},
  year={2018}
}

@Techreport{krizhevsky2009learning,
 author = {Krizhevsky, Alex and Hinton, Geoffrey},
 address = {Toronto, Ontario},
 institution = {University of Toronto},
 number = {0},
 publisher = {Technical report, University of Toronto},
 title = {Learning multiple layers of features from tiny images},
 year = {2009},
 title_with_no_special_chars = {Learning multiple layers of features from tiny images},
 url = {https://www.cs.toronto.edu/~kriz/learning-features-2009-TR.pdf}
}

@inproceedings{grauman2024ego,
  title={Ego-exo4d: Understanding skilled human activity from first-and third-person perspectives},
  author={Grauman, Kristen and Westbury, Andrew and Torresani, Lorenzo and Kitani, Kris and Malik, Jitendra and Afouras, Triantafyllos and Ashutosh, Kumar and Baiyya, Vijay and Bansal, Siddhant and Boote, Bikram and others},
  booktitle={Proc. of CVPR},
  year={2024}
}

@ARTICLE{6702844,
  author={Medjiah, Samir and Taleb, Tarik and Ahmed, Toufik},
  journal={IEEE Transactions on Wireless Communications}, 
  title={Sailing over Data Mules in Delay-Tolerant Networks}, 
  year={2014},
  volume={13},
  number={1},
}

@article{yu2023idml,
  title={idml: Incentivized decentralized machine learning},
  author={Yu, Haoxiang and Chen, Hsiao-Yuan and Lee, Sangsu and Vishwanath, Sriram and Zheng, Xi and Julien, Christine},
  journal={arXiv preprint arXiv:2304.05354},
  year={2023}
}

@article{fallah2020personalized,
  title={Personalized federated learning: A meta-learning approach},
  author={Fallah, Alireza and Mokhtari, Aryan and Ozdaglar, Asuman},
  journal={arXiv preprint arXiv:2002.07948},
  year={2020}
}

@article{hanzely2020federated,
  title={Federated learning of a mixture of global and local models},
  author={Hanzely, Filip and Richt{\'a}rik, Peter},
  journal={arXiv preprint arXiv:2002.05516},
  year={2020}
}

@inproceedings{10.1145/3664647.3681588,
author = {Wu, Xinghao and Liu, Xuefeng and Niu, Jianwei and Wang, Haolin and Tang, Shaojie and Zhu, Guogang and Su, Hao},
title = {Decoupling General and Personalized Knowledge in Federated Learning via Additive and Low-rank Decomposition},
year = {2024},
booktitle = {Proc. of MM},
pages = {7172–7181},
numpages = {10},
}

@article{yi2023pfedes,
  title={pFedES: Model Heterogeneous Personalized Federated Learning with Feature Extractor Sharing},
  author={Yi, Liping and Yu, Han and Wang, Gang and Liu, Xiaoguang},
  journal={arXiv preprint arXiv:2311.06879},
  year={2023}
}

@inproceedings{NEURIPS2020_24389bfe,
 author = {Fallah, Alireza and Mokhtari, Aryan and Ozdaglar, Asuman},
 booktitle = {Advances in Neural Information Processing Systems},
 editor = {H. Larochelle and M. Ranzato and R. Hadsell and M.F. Balcan and H. Lin},
 pages = {3557--3568},
 publisher = {Curran Associates, Inc.},
 title = {Personalized Federated Learning with Theoretical Guarantees: A Model-Agnostic Meta-Learning Approach},
 volume = {33},
 year = {2020}
}

@InProceedings{Lim_2024_WACV,
    author    = {Lim, Jin Hyuk and Ha, SeungBum and Yoon, Sung Whan},
    title     = {MetaVers: Meta-Learned Versatile Representations for Personalized Federated Learning},
    booktitle = {Proc of WACV},
    year      = {2024},
}

@INPROCEEDINGS{9860349,
  author={Kundu, Achintya and Yu, Pengqian and Wynter, Laura and Lim, Shiau Hong},
  booktitle={Proc. of EDGE}, 
  title={Robustness and Personalization in Federated Learning: A Unified Approach via Regularization}, 
  year={2022},
  volume={},
  number={},
}

@InProceedings{10.1007/978-3-031-73110-5_17,
author="Messmer, Liane-Marina
and Reich, Christoph
and Abdeslam, Djaffar Ould",
editor="Arai, Kohei",
title="Context-Aware Machine Learning: A Survey",
booktitle="Proc. of FTC",
year="2024",
pages="252--272",
}

@article{SARKER2020102762,
title = {ABC-RuleMiner: User behavioral rule-based machine learning method for context-aware intelligent services},
journal = {Journal of Network and Computer Applications},
volume = {168},
year = {2020},
author = {Iqbal H. Sarker and A.S.M. Kayes},
}

@article{harries1998extracting,
  title={Extracting hidden context},
  author={Harries, Michael Bonnell and Sammut, Claude and Horn, Kim},
  journal={Machine learning},
  volume={32},
  number={2},
  pages={101--126},
  year={1998},
  publisher={Springer}
}

@misc{Brdiczka_2019, url={https://business.adobe.com/blog/perspectives/contextual-ai-the-next-frontier-of-artificial-intelligence}, journal={Contextual AI: The Next Frontier of Artificial Intelligence}, author={Brdiczka, Oliver}, year={2019}, month={4}}

@article{WANG2023103646,
title = {Context understanding in computer vision: A survey},
journal = {Computer Vision and Image Understanding},
volume = {229},
year = {2023},
author = {Xuan Wang and Zhigang Zhu},
}

@article{miranda2022survey,
  title={A survey on the use of machine learning methods in context-aware middlewares for human activity recognition},
  author={Miranda, Leandro and Viterbo, Jos{\'e} and Bernardini, Fl{\'a}via},
  journal={Artificial Intelligence Review},
  volume={55},
  number={4},
  pages={3369--3400},
  year={2022},
  publisher={Springer}
}

@INPROCEEDINGS{9097597,
  author={Yu, Tianlong and Li, Tian and Sun, Yuqiong and Nanda, Susanta and Smith, Virginia and Sekar, Vyas and Seshan, Srinivasan},
  booktitle={Proc. of IoTDI}, 
  title={Learning Context-Aware Policies from Multiple Smart Homes via Federated Multi-Task Learning}, 
  year={2020},
  volume={},
  number={},
  pages={104-115},
}

@ARTICLE{9931527,
  author={Huang, Guang-Li and Zaslavsky, Arkady and Loke, Seng W. and Abkenar, Amin and Medvedev, Alexey and Hassani, Alireza},
  journal={IEEE Transactions on Intelligent Transportation Systems}, 
  title={Context-Aware Machine Learning for Intelligent Transportation Systems: A Survey}, 
  year={2023},
  volume={24},
  number={1},
  pages={17-36},
}

@inproceedings{nascimento2018context,
  title={A context-aware machine learning-based approach},
  author={Nascimento, Nathalia and Alencar, Paulo and Lucena, Carlos and Cowan, Donald},
  booktitle={Proceedings of the 28th Annual International Conference on Computer Science and Software Engineering},
  pages={40--47},
  year={2018}
}

@article{sim2018online,
  title={An online context-aware machine learning algorithm for 5G mmWave vehicular communications},
  author={Sim, Gek Hong and Klos, Sabrina and Asadi, Arash and Klein, Anja and Hollick, Matthias},
  journal={IEEE/ACM Transactions on Networking},
  volume={26},
  number={6},
  pages={2487--2500},
  year={2018},
  publisher={IEEE}
}

@article{liu2017context,
  title={Context aware machine learning approaches for modeling elastic localization in three-dimensional composite microstructures},
  author={Liu, Ruoqian and Yabansu, Yuksel C and Yang, Zijiang and Choudhary, Alok N and Kalidindi, Surya R and Agrawal, Ankit},
  journal={Integrating Materials and Manufacturing Innovation},
  year={2017},
  publisher={Springer}
}

@misc{menik2023modularmachinelearningsolution,
      title={Towards Modular Machine Learning Solution Development: Benefits and Trade-offs}, 
      author={Samiyuru Menik and Lakshmish Ramaswamy},
      year={2023},
      eprint={2301.09753},
      archivePrefix={arXiv},
      primaryClass={cs.LG},
      url={https://arxiv.org/abs/2301.09753}, 
}

@InProceedings{Sarker2021,
author="Sarker, Iqbal H.
and Colman, Alan
and Han, Jun
and Watters, Paul",
title="Introduction to Context-Aware Machine Learning and Mobile Data Analytics",
bookTitle="Context-Aware Machine Learning and Mobile Data Analytics: Automated Rule-based Services with Intelligent Decision-Making",
year="2021",
pages="3--13",
}

@article{BAYOUDH2024102217,
title = {A survey of multimodal hybrid deep learning for computer vision: Architectures, applications, trends, and challenges},
journal = {Information Fusion},
volume = {105},
year = {2024},
author = {Khaled Bayoudh},
}

@Article{app12189305,
AUTHOR = {Omolaja, Adebola and Otebolaku, Abayomi and Alfoudi, Ali},
TITLE = {Context-Aware Complex Human Activity Recognition Using Hybrid Deep Learning Models},
JOURNAL = {Applied Sciences},
VOLUME = {12},
YEAR = {2022},
NUMBER = {18},
ARTICLE-NUMBER = {9305},
}

@article{BEJANI2018303,
title = {A context aware system for driving style evaluation by an ensemble learning on smartphone sensors data},
journal = {Transportation Research Part C: Emerging Technologies},
volume = {89},
pages = {303-320},
year = {2018},
author = {Mohammad Mahdi Bejani and Mehdi Ghatee}
}

@article{WU2022179,
title = {Knowledge graph-based multi-context-aware recommendation algorithm},
journal = {Information Sciences},
volume = {595},
pages = {179-194},
year = {2022},
issn = {0020-0255},
url = {https://www.sciencedirect.com/science/article/pii/S0020025522001967},
author = {Chao Wu and Sannyuya Liu and Zeyu Zeng and Mao Chen and Adi Alhudhaif and Xiangyang Tang and Fayadh Alenezi and Norah Alnaim and Xicheng Peng}
}

@INPROCEEDINGS{10298476,
  author={King, Evan and Julien, Christine},
  booktitle={Proc. of MASS}, 
  title={CANDor: Continuous Adaptive Neighbor Discovery}, 
  year={2023},
  volume={},
  number={},
  pages={336-342},
}

@InProceedings{pmlr-v54-mcmahan17a,
  title = 	 {{Communication-Efficient Learning of Deep Networks from Decentralized Data}},
  author = 	 {McMahan, Brendan and Moore, Eider and Ramage, Daniel and Hampson, Seth and Arcas, Blaise Aguera y},
  booktitle = 	 {Proc. of AISTATS},
  pages = 	 {1273--1282},
  year = 	 {2017},
  editor = 	 {Singh, Aarti and Zhu, Jerry},
  volume = 	 {54},
  month = 	 {4},
}

@inproceedings{acar2021federated,
title={Federated Learning Based on Dynamic Regularization},
author={Durmus Alp Emre Acar and Yue Zhao and Ramon Matas and Matthew Mattina and Paul Whatmough and Venkatesh Saligrama},
booktitle={Proc. of ICLR},
year={2021},
}

@InProceedings{pmlr-v119-karimireddy20a,
  title = 	 {{SCAFFOLD}: Stochastic Controlled Averaging for Federated Learning},
  author =       {Karimireddy, Sai Praneeth and Kale, Satyen and Mohri, Mehryar and Reddi, Sashank and Stich, Sebastian and Suresh, Ananda Theertha},
  booktitle = 	 {Proc. of ICML},
  pages = 	 {5132--5143},
  year = 	 {2020},
}

@misc{li2020federatedoptimizationheterogeneousnetworks,
      title={Federated Optimization in Heterogeneous Networks}, 
      author={Tian Li and Anit Kumar Sahu and Manzil Zaheer and Maziar Sanjabi and Ameet Talwalkar and Virginia Smith},
      year={2020},
      eprint={1812.06127},
      archivePrefix={arXiv},
      primaryClass={cs.LG},
      url={https://arxiv.org/abs/1812.06127}, 
}

@inproceedings{mcmahan2017communication,
  title={Communication-efficient learning of deep networks from decentralized data},
  author={McMahan, Brendan and Moore, Eider and Ramage, Daniel and Hampson, Seth and y Arcas, Blaise Aguera},
  booktitle={Artificial intelligence and statistics},
  pages={1273--1282},
  year={2017},
  organization={PMLR}
}

@InProceedings{Yang_2024_CVPR,
    author    = {Yang, Xiyuan and Huang, Wenke and Ye, Mang},
    title     = {FedAS: Bridging Inconsistency in Personalized Federated Learning},
    booktitle = {Proc. of CVPR},
    month     = {6},
    year      = {2024},
    pages     = {11986-11995}
}

@misc{sattler2019clusteredfederatedlearningmodelagnostic,
      title={Clustered Federated Learning: Model-Agnostic Distributed Multi-Task Optimization under Privacy Constraints}, 
      author={Felix Sattler and Klaus-Robert Müller and Wojciech Samek},
      year={2019},
      eprint={1910.01991},
      archivePrefix={arXiv},
      primaryClass={cs.LG},
      url={https://arxiv.org/abs/1910.01991}, 
}

@misc{hsu2019measuringeffectsnonidenticaldata,
      title={Measuring the Effects of Non-Identical Data Distribution for Federated Visual Classification}, 
      author={Tzu-Ming Harry Hsu and Hang Qi and Matthew Brown},
      year={2019},
      eprint={1909.06335},
      archivePrefix={arXiv},
      primaryClass={cs.LG},
      url={https://arxiv.org/abs/1909.06335}, 
}

@ARTICLE{9043535,
  author={Xia, Kun and Huang, Jianguang and Wang, Hanyu},
  journal={IEEE Access}, 
  title={LSTM-CNN Architecture for Human Activity Recognition}, 
  year={2020},
  volume={8},
  number={},
}

@misc{Foursquare,
  title = {Foursquare},
  howpublished = {\url{https://foursquare.com/products/visits/}}
}

@misc{jetson,
  title = {Jetson Orin Nano},
  howpublished = {\url{https://www.nvidia.com/en-us/autonomous-machines/embedded-systems/jetson-orin/}}
}

@misc{rp5,
  title = {Raspberry Pi 5},
  howpublished = {\url{https://www.raspberrypi.com/products/raspberry-pi-5/}}
}

@article{izmailov2018averaging,
  title={Averaging weights leads to wider optima and better generalization},
  author={Izmailov, Pavel and Podoprikhin, Dmitrii and Garipov, Timur and Vetrov, Dmitry and Wilson, Andrew Gordon},
  journal={arXiv preprint arXiv:1803.05407},
  year={2018}
}

@article{lei2017insecurity,
  title={The insecurity of home digital voice assistants--amazon alexa as a case study},
  author={Lei, Xinyu and Tu, Guan-Hua and Liu, Alex X and Ali, Kamran and Li, Chi-Yu and Xie, Tian},
  journal={arXiv preprint arXiv:1712.03327},
  year={2017}
}

@inproceedings{cho2020will,
  title={Will deleting history make alexa more trustworthy? effects of privacy and content customization on user experience of smart speakers},
  author={Cho, Eugene and Sundar, S Shyam and Abdullah, Saeed and Motalebi, Nasim},
  booktitle={Proceedings of the 2020 CHI conference on human factors in computing systems},
  pages={1--13},
  year={2020}
}

@article{huang2024federated,
  title={Federated learning-empowered AI-generated content in wireless networks},
  author={Huang, Xumin and Li, Peichun and Du, Hongyang and Kang, Jiawen and Niyato, Dusit and Kim, Dong In and Wu, Yuan},
  journal={IEEE Network},
  volume={38},
  number={5},
  pages={304--313},
  year={2024},
  publisher={IEEE}
}
